\pdfoutput=1
\documentclass[conference]{IEEEtran}
\usepackage[numbers]{natbib}    
\usepackage{amsmath}            
\usepackage{amssymb}            
\usepackage{cleveref}           
\usepackage{graphicx}
\usepackage{duckuments}         
\usepackage{algorithm}
\usepackage{algpseudocode}      
\usepackage{booktabs}
\usepackage{multirow, makecell}
\usepackage[hyphens]{url}       
\usepackage[version=4]{mhchem}  

\usepackage[mode=buildnew]{standalone}
\usepackage{tikz}
\usetikzlibrary{positioning,fit}
\tikzset{
    block/.style={
        draw,
        minimum width=50, 
        minimum height=30,
    },
    items/.style={
        font=\footnotesize,
    },
}

\usepackage{soul}

\crefname{paragraph}{paragraph}{paragraphs}
\Crefname{paragraph}{Paragraph}{Paragraphs}

\DeclareMathOperator*{\informer}{\mathcal{I}}
\DeclareMathOperator*{\encoder}{\mathcal{E}}
\DeclareMathOperator*{\classifier}{\mathcal{C}}
\DeclareMathOperator*{\observer}{\mathfrak{O}}
\DeclareMathOperator*{\threatmodel}{\mathfrak{T}}
\DeclareMathOperator*{\blackbox}{\mathfrak{G}}
\DeclareMathOperator*{\graybox}{\mathfrak{B}}
\DeclareMathOperator*{\distribution}{\mathcal{D}}
\DeclareMathOperator*{\append}{\mathbin\Vert}

\title{Privacy Threats in Stable Diffusion Models}
\makeatletter
\author{
\IEEEauthorblockN{Thomas Cilloni}
\IEEEauthorblockA{%
    \textit{University of Mississippi}\\
    Oxford, MS \\
    tcilloni@go.olemiss.edu}%
\and
\IEEEauthorblockN{Charles Fleming}
\IEEEauthorblockA{%
    \textit{Cisco}\\
    San Jose, CA \\
    chflemin@cisco.com}%
\and
\IEEEauthorblockN{Charles Walter}
\IEEEauthorblockA{%
    \textit{University of Mississippi}\\
    Oxford, MS \\
    cwwalter@olemiss.edu}%
}%

\begin{document}

\maketitle

\begin{abstract}
    This paper introduces a novel approach to membership inference attacks (MIA) targeting stable diffusion computer vision models, specifically focusing on the highly sophisticated Stable Diffusion V2 by StabilityAI. MIAs aim to extract sensitive information about a model's training data, posing significant privacy concerns. 
    Despite its advancements in image synthesis, our research reveals privacy vulnerabilities in the stable diffusion models' outputs. Exploiting this information, we devise a black-box MIA that only needs to query the victim model repeatedly.
    Our methodology involves observing the output of a stable diffusion model at different generative epochs and training a classification model to distinguish when a series of intermediates originated from a training sample or not. We propose numerous ways to measure the membership features and discuss what works best. The attack's efficacy is assessed using the ROC AUC method, demonstrating a 60\% success rate in inferring membership information.     
    This paper contributes to the growing body of research on privacy and security in machine learning, highlighting the need for robust defenses against MIAs. Our findings prompt a reevaluation of the privacy implications of stable diffusion models, urging practitioners and developers to implement enhanced security measures to safeguard against such attacks.
\end{abstract}


\section{Introduction}
Machine learning (ML) has recently improved tremendously, achieving extraordinary results in countless learning tasks, including object recognition \cite{object_recognition}, natural language processing \cite{bert}, and data generation \cite{covidgan}, as well as advanced applications such as healthcare analysis \cite{healthcare_ai} and self-driving intelligence \cite{self_driving}. Two major enabling factors make such advancements possible: improved hardware computational abilities and availability of larger datasets \cite{rise_of_ai}.

Several major security and privacy threats exist in the context of machine learning. As \citet{security_and_privacy} explain, potential attackers of ML systems can target their integrity, such as by inducing errors in face recognition systems \cite{fawkes,ulixes}, or their availability, simply with DDoS attacks \cite{security_and_privacy}, but also their confidentiality, particularly in financial marketing systems or medical models \cite{security_and_privacy_33,security_and_privacy_34}, and their privacy, which we amply discuss here. One primary concern with privacy-preserving ML practitioners is the potential exposure of sensitive or confidential information present in a training dataset. As datasets can contain individuals’ private information, such as user speech, images, and medical records, it is essential that ML models should not leak privacy-sensitive information about their training data. This could include personally identifiable information (PII), proprietary data, or any details that, when disclosed, could compromise privacy, security, or business interests. Another risk is the possibility of re-identification of individuals in the training dataset. Even if explicit personal details are not disclosed, the combination of seemingly innocuous data points may allow adversaries to infer missing attributes from those available and make it easier to link records back to specific individuals. 

Furthermore, unintended information leakage can lead to unauthorized insights into the characteristics and distribution of the training data. This may empower malicious actors to deduce patterns, vulnerabilities, or biases in the training set, undermining the model's fairness, accountability, and trust in users. Overall, privacy risks associated with model information leakage highlight the need for robust mechanisms to ensure that machine learning models do not inadvertently expose sensitive or confidential data during their operation.
\begin{figure}
    \centering
    \includegraphics[width=\columnwidth]{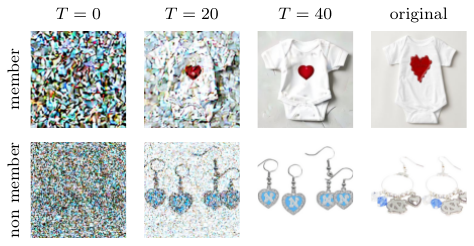}
    \caption{Examples of images being generated by Stable Diffusion V2 from a member and a nonmember images. The image generated on top is strikingly similar to its corresponding sample from the training set, whereas the one on the bottom is noticeably different from what was seen at train time.}
    \label{fig:diffusion_examples_small}
\end{figure}

Recent research has underscored the susceptibility of machine learning (ML) models to memorizing details from their training data \cite{loss}, rendering them susceptible to a range of privacy threats, including but not limited to model extraction attacks\cite{model_extraction}, attribute inference attacks (also referred to as model inversion attacks)\cite{attribute_inference,model_inversion}, property inference attacks\cite{property_inference}, and membership inference attacks \cite{loss,lira}. Going from least to most privacy violating: model extraction attacks try to replicate a target model, thereby exposing its architecture and learned parameters, which is nowadays considered equivalent to exposing its underlying training data entirely \cite{models_as_data}; attribute inference attacks seek to deduce sensitive attributes of a given data record based on model outputs and non-sensitive attribute information; property inference attacks are similar, with the difference that the latter focus on populations and not individual records; finally, membership inference attacks (MIAs) aim to discern the members constituting the training dataset of an ML model. MIAs might be the least concerning from a privacy perspective compared to others. However, two key features make them equally concerning and most widely used for privacy risk assessments \cite{models_as_data}: they are the easiest/strongest attacks to carry out, as they try to infer the least amount of information (a single bit per training sample, to be exact), and they can be used as the foundation for more privacy-invasive attacks \cite{MIA_FP_3,MIA_FP_4}.

Membership Inference Attacks (MIAs) on Machine Learning (ML) models are used to determine whether a data record was used to train a target ML model or not. Such attacks can pose significant privacy risks to individuals. For instance, if a clinical record is used to train an ML model associated with a certain disease, MIAs can infer that the owner of the clinical record has that disease with a high probability. The National Institute of Standards and Technology (NIST) recently published a report\cite{nist} that specifically mentions that a MIA that determines an individual's inclusion in the dataset used to train the target model violates confidentiality. Furthermore, MIAs can lead to commercial companies violating privacy regulations, especially those that provide Machine Learning as a Service (MLaaS). MIAs on ML models can increase the risks of such models being classified as private personal information under the General Data Protection Regulation (GDPR)\cite{GDPR}.

The concept of MIAs was first proposed by \citet{MIA_FP_22}, where they demonstrated how an attacker could leverage published statistics about a genomics dataset to infer the presence of a particular genome in the dataset. In the context of ML, \citet{MIA_FP_60} proposed the first MIAs on several classification models. They demonstrated that an attacker can identify whether a data record was used to train a neural network-based classifier or not solely based on the prediction vector of the data record (also known as black-box access to the target ML model). Since then, there has been an increasing number of studies that investigate MIAs on various ML models, including regression models\cite{mia_regression}, classification models\cite{lira}, generation models\cite{MIA_vs_SD_USENIX}, and embedding models\cite{mia_embedding}. Meanwhile, several studies\cite{memguard,dp,darknetz} propose different membership inference defenses from different perspectives to defend against MIAs while preserving the utility of the target ML models.

This paper proposes a novel approach to membership inference attacks that applies to generative AI models and demonstrates its effectiveness on StabilityAI's Stable Diffusion V2. Traditional MIAs exploit the fact that models tend to memorize training data samples and, therefore, exhibit distinctive behaviors when fed them. Such observation cannot be made with generative models in a strictly black-box setting, as other models cannot accurately estimate ground truth labels. To overcome this challenge, we propose a method that instead compares a fed image into a diffusion model and the output(s) it produces, guided by a textual prompt. We explore several flavors of this method, using:
\begin{itemize}
    \item Different access modes to the victim model, leading to two \textit{gray-box} attacks and one strictly \textit{black-box} attack. These are presented in detail in \Cref{sec:observer}.
    \item An optional smoothing of the generated outputs of the victim model, to weed out noise and imperfections in images, to try to boost the attacks' performance. Relevant results are in \Cref{sec:results}.
    \item Several visual comparison metrics to measure the difference between expected and actual outputs of the target models. These include three visual-based comparison metrics (namely PSNR, DSSIM, and RMSE), and a feature-based comparison metric that measures the similarity between the feature vectors that a pre-trained ResNet-50 model extracts from images. See \Cref{sec:distance_metrics} for a detailed discussion.
\end{itemize}

This method is the first successful attempt at membership inference against diffusion models in black-box settings. Our results show that the membership status of training data points from a balanced dataset can be inferred with 60\% accuracy (measured with the ROC AUC method) from an industry-grade stable diffusion model.


\section{Background}
Here, we briefly introduce basic notions of membership inference attacks and stable diffusion generative image models. 

\subsection{Data Privacy}
The field of data privacy is concerned with studying attacks that leak information about data, in this case, the training data of a machine learning model. The attacks are studied both from an adversarial perspective, trying to maximize their effectiveness, and from a defender perspective, developing ways to measure their impact and researching prevention techniques.

\subsubsection{Privacy Attacks}
Privacy Attacks are hostile attacks against machine learning models that try to extract information about their training data. The most widely recognized types of privacy attacks are \textit{data extraction}, where an adversary tries to recover specific training data points \cite{MIA_vs_SD_USENIX} (e.g., text messages from a language model); \textit{model inversion}, where aggregate information is instead extracted, usually about a certain sub-population of the training dataset \cite{attribute_inference,model_inversion} (e.g., chances of credit card approval by ZIP code); \textit{property inference}, in which non-trivial properties of training samples are attempted to be extracted (e.g., determining whether Bitcoin log machines are patched for Spectre and Meltdown \cite{MIA_FP_17}). This paper focuses on \textit{membership inference} attacks. These were first studied in the context of medical data, where determining the presence of a medical record of a patient in a dataset was by itself a privacy violation \cite{MIA_FP_11,MIA_FP_12,MIA_FP_22,MIA_FP_59}. Membership inference attacks are particularly important because they allow stronger extraction attacks to be executed \cite{MIA_FP_3,MIA_FP_4}.

\subsubsection{Data Memorization}
The ability of a model to perform membership inference depends on its competence to recall individual data points or labels. Zhang et al. \cite{MIA_FP_72} observed that standard neural networks can memorize datasets with randomly assigned labels. Furthermore, \citet{MIA_FP_14}, among others \cite{MIA_FP_2,MIA_FP_15}, has conducted extensive theoretical and empirical research, which has shown that some level of memorization may be necessary to achieve optimal generalization. As such, it can be inferred that the capacity of a model to remember data points and labels is a crucial factor in determining its ability to carry out membership inference.

\subsubsection{Defense Mechanisms}
Several training techniques exist to prevent privacy attacks, and we refer to these under the umbrella term of \textit{privacy-preserving training}. In contrast with many security-based attacks, while a privacy-preserving training technique may be proven effective in a set of use cases \cite{MIA_FP_26,MIA_FP_45}, its effectiveness cannot be formally demonstrated \cite{MIA_FP_6,MIA_FP_61}. The only notable exception is introducing differential privacy in a model's training \cite{MIA_FP_10}, such as by model ensembling \cite{MIA_FP_50} or by making the optimization algorithm differentially private \cite{MIA_FP_1,MIA_FP_64}. Differential privacy, however, comes with a significant privacy-accuracy tradeoff \cite{dp_cost}.

\subsection{Membership Inference Attacks}
The goal of a membership inference attack (MIA) is to recognize what data samples were used in training a model, ignoring those that were not. MI attacks are the most fundamental attack on data privacy and are widely used for measuring the privacy of a training dataset. This section formalizes MI attacks as a security game and briefly describes common evaluation metrics.

\subsubsection{Membership Inference Attacks Formalized}
\label{sec:security_game}
The definition of a membership inference attack as a security game follows from \cite{loss} and \cite{MIA_FP_25}. The game has an attacker who is the carrier of the membership inference attack and a defender who is the owner of the victim ML model $f$ and has knowledge of its training data distribution $\distribution$.
\begin{enumerate}
    \item The defender samples a training dataset $D_\text{train} \gets \distribution$, and optionally additional testing/validation datasets $D^* \gets \distribution$, and trains a model $f_\theta \gets \mathcal{T}(D_\text{train})$.
    \item The defender flips a coin $c_D$, and if the result is heads, gives the adversary a sample $(x,y) \in D_\text{train}$; otherwise, it supplies a sample $(x,y) \in \distribution / D_\text{train}$.
    \item The adversary, with oracle access to the model $f_\theta$ and the distribution $\distribution$, tries to guess whether $(x,y)$ is originated from $D_\text{train}$ or from $\distribution / D_\text{train}$. Their guess is produced using a predictive model/system $M$ that can use both $\distribution$ and $f_\theta$, and is registered as a coin face $c_A \gets M(x, y, \distribution, f_\theta)$.
    \item If the adversary guesses the coin face correctly ($c_A = c_D$), the membership of the sample is correctly inferred. Over multiple iterations of the game, one can measure the attacker's power.
\end{enumerate}

In this formulation, the attacker is given both samples from the training data distribution and their associated labels. The attacker also has query access to the training data distribution, which allows them to train a virtually unlimited number of shadow models to mimic the behavior of the victim model \cite{MIA_FP_60}, thereby enabling theoretically more powerful attacks. These two assumptions are what MI attacks commonly require, and this threat model is most widely used in the literature; however, an attacker does not necessarily need to use all of this information to carry out MI attacks.

\subsubsection{Evaluating Membership Inference Attacks}
\label{sec:background_evaluation}
The security game proposed above works well in the individual sample case, but when the membership of several samples is to be inferred, we can use a more sophisticated evaluation method. An attacker is rarely 100\% 

MI attacks are evaluated as a standard binary classifier using the AUC-ROC curve method. The Receiver Operating Characteristics (ROC) curve is a plot of the True Positive Rate (TPR) against the False Positive Rate (FPR) of a binary classifier at increasing classification thresholds. For simplicity, features are within the 0-1 range, and so are the thresholds. The Area Under the Curve (AUC) is the area under the ROC curve in a 0-1 range and falls in the 0-1 range. Higher AUC-ROC values indicate better performance, and a value of 0.5 is the worst case, equivalent to the random toss of a coin. A good binary classifier can raise its AUC-ROC by increasing its TPR at any FPR value. Recent work \cite{lira} suggests that visual ROC curves should always accompany their resulting AUC-ROC results and should be plotted both on a linear scale and a logarithmic scale.

Finally, we reconsider the MI model $M$ to be a sequential pipeline, whose first part $E$ is an encoder that produces a set of $N$ features for a given data sample, and a second part is a binary classification mechanism $B$ that processes those features to produce a membership confidence score $c$, as $M = B \circ E$.

\subsubsection{Notable Examples of Membership Inference Attacks}
One of the most notable examples of membership inference attacks is the work by \citet{MIA_FP_60}. This is the first time a \textit{shadow model training} procedure is introduced. Shadow training consists of training one or several machine learning models to mimic the behavior of another model to attack, the victim model, to which the attacker only has black-box oracle access. This procedure aims to obtain one or more models that behave like the model to attack and of which the attacker knows the training dataset. This, in turn, allows the attacker to train a binary classification model on the outputs of the shadow models since the ground truth labels are, in this case, available. The model is then used to infer the membership of the samples supplied by the defender to the training dataset of the victim model. \citet{MIA_FP_60} propose three techniques to train the shadow models: the first method uses only black-box access to the victim model, mimicking its outputs; the second uses statistics about the population of the training dataset distribution, therefore allowing the attacker to construct similar training datasets for their shadow models; the last gives the attacker access to a noisy version of the victim model's training dataset. The results show that membership inference can reach accuracies of 70\% to 90\% for small image datasets and census/medical records datasets, highlighting privacy vulnerabilities in commercial services from multiple tech giants, among others.

An evolution of the MI attack in \cite{MIA_FP_60} is the label-only MI attack proposed by \citet{MIA_FP_6}. While previous attacks to that date all used the entirety of a model's output (e.g., a probability distribution over the possible labels of a classifier), label-only attacks only require the final labels (e.g., the most confident class in a classifier's output), which is post-processed output. This attack was quite revolutionary as several defense mechanisms against MI attacks were proposed at the time, all revolving around the idea of hiding the confidence of the predictions \cite{memguard,MIA_FP_60} to prevent guessing by error or training shadow models effectively. Label-only attacks make repeated predictions on a single sample that is increasingly perturbed until pushed beyond its originally predicted class onto another, therefore giving a measure of the sample's distance to the victim model's decision boundary for that class. This distance can give away whether a sample was used for training or not: the deeper a sample is within a class' decision boundary, the more likely it was seen during training.

One of the most recent and notable examples of MI attacks is that of LiRA \cite{lira}. The LiRA attack combines several ideas from the literature on MI attacks to produce a model that is as effective as any prior work on existing metrics. But perhaps the most notable contribution of this paper is the new evaluation approach that the authors propose. They argue that many membership inference attacks are non-membership inference attacks because they excel at picking out non-members but struggle, if not fail, at picking out members. Intuitively, an MI attack built around the concept of memorization can confidently tell that a sample was not seen during training if the error of the victim model on it is high (or equivalently if its confidence is low); however, if the error on a sample prediction is low, there is no guarantee that the sample is indeed from the training set. \citep{lira} show an example of such a model and suggest using log-scaled ROC curves to show the attacker's power in a membership inference attack.

\subsection{Stable Diffusion in Computer Vision}
Denoising diffusion models has recently surfaced as a cutting-edge field within computer vision, exhibiting impressive outcomes in the realm of generative modeling. These models delve into the depths of generative modeling through a unique process involving two key phases: forward diffusion and reverse diffusion. The initial data is systematically modified over multiple iterations during the forward diffusion phase by introducing Gaussian noise. Subsequently, in the reverse diffusion stage, the model endeavors to painstakingly reverse the perturbations step by step, ultimately restoring the original input data. Despite their acknowledged computational challenges, primarily stemming from the substantial number of steps required during sampling, diffusion models are highly praised for the excellence and variety of the samples they generate.

Up to this point, diffusion models have proven themselves particularly versatile, with applications spanning a wide spectrum of generative modeling tasks. Whether it is in image generation 
\cite{SD_survey_2,SD_survey_3}, image super-resolution \cite{SD_survey_10,SD_survey_23}, image inpainting \cite{SD_survey_3,SD_survey_4}, or image editing \cite{SD_survey_31,SD_survey_32}, diffusion models have achieved remarkable results. In addition, these models have shown to be capable of tasks beyond purely generative tasks. The latent representations they learn are applicable to discriminative tasks such as image segmentation \cite{SD_survey_39,SD_survey_40}, classification \cite{SD_survey_43}, and anomaly detection \cite{SD_survey_44,SD_survey_45}. This rich array of applications highlights the extensive reach of denoising diffusion models, suggesting that the horizons of their potential uses are yet to be exhaustively explored.

\subsubsection{Stable Diffusion Pre-DDPM}
Diffusion models (DMs), also known as diffusion probabilistic models \cite{tti_29}, are generative models built as Markov chains trained with variational inference \cite{SD_survey_2}. Their core goal is to perturb data through diffusion for sample generation \cite{tti_29}, \cite{SD_survey_2}. A significant milestone, the denoising diffusion probabilistic model (DDPM) \cite{SD_survey_2}, was introduced in 2020, sparking widespread interest in the generative modeling community.

The emergence of the new state-of-the-art Denoising Diffusion Probabilistic Models (DDPM) \cite{SD_survey_2} can be largely traced back to two pioneering approaches: Score-Based Generative Models (SGM) \cite{SD_survey_3}, explored in 2019, and Diffusion Probabilistic Models (DPM) \cite{tti_29}, which made their debut as early as 2015. To understand DDPM comprehensively, it is essential to revisit the foundational principles of DPM \cite{tti_29} and Score-Based Generative Models \cite{SD_survey_3} (SDE) before delving into DDPM.

Diffusion Probabilistic Models (DPM) \cite{tti_29} are groundbreaking in modeling probability distributions by estimating the inverse of a Markov diffusion chain that simplifies complex data distributions. DPM \cite{tti_29} uses a forward process to simplify complex data and then learns to reverse this diffusion process, as evidenced by empirical results on various datasets. DDPM \cite{SD_survey_2} further refines and optimizes DPM \cite{tti_29} to enhance its implementations.

Score-Based Generative Models (SGM) \cite{SD_survey_3} play a significant role here. SGM explores techniques to enhance generative models in \cite{SD_survey_3} and \cite{tti_32}. SGM perturbs data with varying Gaussian noise, uses the gradient of log probability density as the score function to generate samples with reduced noise levels, and estimates score functions for noisy data distributions during training. Despite different motivations, SGM and DDPM share a common optimization goal during training, as discussed in \cite{SD_survey_2}. Additionally, \cite{tti_32} delves into an improved SGM version for high-resolution image generalization.

\subsubsection{Denoising Diffusion Probabilistic Models}
DDPMs, or denoising diffusion probabilistic models, are a type of Markov chain that can create images from noise through a limited number of transitions during inference. These transitions are learned during training by perturbing natural images with noise in a reversed direction, with each step adding noise to the data and optimizing it as the target.

The forward pass of a denoising diffusion probabilistic model takes in a noisy image as input and applies a series of transformations, known as transitions, to the image. Each transition adds more noise to the image, making it more difficult to observe the true underlying image. However, the transitions are carefully designed such that the noise added at each step follows a specific probabilistic distribution that can be modeled and learned during training. 

As the model progresses through the transitions, it gradually removes the added noise and reveals the true underlying image. This process is repeated for a fixed number of steps until the image is completely denoised. The output of the forward pass is a denoised image that estimates the true underlying image that generated the noisy input.

In the reverse pass, the model takes the denoised image that was output by the forward pass and generates a sequence of noise levels that, when added to the denoised image, produce the original noisy input. This is achieved by starting with the denoised image and applying the transitions in the reverse order, subtracting the learned noise distribution at each step to arrive at the original noisy input. 

The reverse pass is useful because it allows for the model to be trained using a maximum likelihood estimation approach, where the goal is to find the noise distribution that is most likely to produce the noisy input given the denoised output. By using the reverse pass to generate the noise sequence, the model can be trained to minimize the difference between the noisy input and the denoised output, maximizing the likelihood of the noise distribution that produced the noisy input.


\section{Related Works}
Stable diffusion models are a very recently developed set of models and techniques in computer vision. As such, their privacy and security have been studied only limitedly. To the best of our knowledge, the works of \citet{MIA_vs_SD_USENIX} and \citet{MIA_vs_SD_ACM} are the only currently peer-reviewed research artifacts that study membership inference attacks against stable diffusion vision models. We discuss these works here in detail. For the sake of completeness, we also briefly discuss two pre-print articles, even if they are currently non-peer-reviewed.

\subsubsection{Extraction and Filter MI Attack}
\citet{MIA_vs_SD_USENIX} have recently successfully transferred the LiRA \cite{lira} and the LOSS \cite{loss} membership inference attacks to stable diffusion models and proposed a new MI attack altogether. This research aims to infer the membership of certain examples and extract them from the training dataset. The extraction is carried out with what the authors describe as an extraction-and-filter pipeline, which uses the victim stable diffusion model to generate several samples from the same text prompt but different random seeds; these are then compared against each other, and if a group of images from the same prompt is visually similar, the stable diffusion model is likely to have seen that image/text pair at training time. The extracted images expose the membership status of certain training samples. Because the model outputs full images, the training samples themselves are extracted, which is a data extraction attack. As we are interested in the MI attack part of this study, that is what the rest of this discussion focuses on.

The membership inference attack implemented by \citet{MIA_vs_SD_USENIX} is a black-box method that only needs to query the victim model, in this case, a Stable Diffusion model trained on 160M images. The attack first requires a careful selection of images from the known training data distribution that have been duplicated repeatedly in the train set: this implies that the victim model has memorized them better, making the membership attack easier. The authors pick 350,000 most-duplicated images, and for each one, query the victim model 500 times to produce a set of candidate training samples. Each sample from the group of 500 images is then compared against one another to find their Euclidean distance, and the images are conceptually laid out in a connected graph. It then follows that a memorized training sample is found if an edge in the graph is short enough. To minimize false positives, however, \citet{MIA_vs_SD_USENIX} pick out only samples that form at least a 10-clique in the graph, following manual hyperparameter tuning. Of the 175M generated images, 94 groups are found to have 10-cliques, and an equal number of images is extracted (to which the authors add 13 extra manually selected samples after visually inspecting the 1000 most dense clusters). The first 50 images are successfully extracted with no false positives, and overall, the 94 images are extracted with a precision above 50\% (though the exact number is not specified). This remarkable result shows that a carefully executed attack can confidently extract training images, even if only very few (0.027\% of the training set) before the precision drops.

The authors also explore how two white-box attacks, namely the LOSS and the LiRA attacks, can be applied to stable diffusion models. \citet{loss} first introduced the loss threshold (LOSS) attack as a simple classifier based on the error of a model. Assuming that there is a minimum generalization gap between train and test performance, a model's error on training samples should be lower than on non-training samples. \citet{MIA_vs_SD_USENIX} use the same loss function that stable diffusion models are trained with to measure the error and get a 0.613 AUC-ROC on a small diffusion model trained on the CIFAR-10 dataset. Similarly, the LiRA attack proposed by \citet{lira} is adapted to stable diffusion models. The attack consists of sampling several shadow training datasets from the given training data distribution and training as many shadow models as possible in the same way the victim stable diffusion model is trained. In their experiments, 16 shadow models are trained, and then the errors they produce when processing training and non-training samples are collected in two sets: IN for members and OUT for non-members. Finally, a binary classifier is trained to distinguish losses from the IN set from those in the OUT set, and the classifier is then used as a membership inference model on the losses of the victim model. In this case, a small stable diffusion model trained on the CIFAR-10 dataset shows a vulnerability of 0.982 with the AUC-ROC method and 0.997 if the attack is augmented with a noise-resistant loss and augmented images.

\subsubsection{Error-based MI Attack}
The second significant work in the field is the research by \citet{MIA_vs_SD_ACM}, who propose a white-box membership inference attack against stable diffusion models. The attack is a straightforward adaptation of the LOSS attack by \citet{loss} and uses the victim model's error on predictions to determine membership. This attack is based on the assumption that models tend to memorize training samples to a certain extent and, therefore, show lower prediction errors. Experiments are carried out on toy models trained on CIFAR-10/100, STL10-U, and Tiny-IN and show that attacks almost always achieve a 0.8 AUC-ROC score and register remarkably high true positive rates (TPR) at low and fixed false positive rates (FPR), in some instances exceeding 30\% TPR@FPR 1\%. While the authors argue that the attack follows a black-box setting, in practice, giving the attacker access to the intermediate diffusions at different timesteps is a gray-box attack, and access to the error of the victim model makes this attack unequivocally white-box.

\subsubsection{Non-Peer Reviewed Work}
Two other recent works are a series of white, gray, and black-box MI attacks \cite{pokemon}, and a white-box and a black-box attack in \cite{japan}. 
\cite{pokemon} propose a series of MI attacks based on notions of errors in stable diffusion models. The authors identify three such metrics, namely the loss function (which makes the attacks white-box), pixel error between the generated and the expected images (which the authors also apply in a black-box setting), and latent error, which is applied in both a white-box and a gray-box setting: if grey-box, an image is first generated from the text prompt, and then the latent representations of the generated image and the original image are compared. Results show that the attack is effective even in the black-box setting with an extremely small dataset (the POKEMON dataset, split into 433 training and 200 testing images). However, a larger dataset makes the black-box and the gray-box implementations both ineffective, with $<$1\% TPR@FPR 1\% (worse than random guess). Additionally, but very importantly, it must be noted that \cite{pokemon} violates the security game described in Section \ref{sec:security_game}: the members and non-member samples from the LAION dataset are sampled from the LAION-Aesthetics V2 5+ and the LAION 2B Multi, respectively; the victim model, Stable Diffusion V1.4, is fine-tuned on the LAION-Aesthetics V2 5+, but from a checkpoint that was previously trained on a subset of LAION 2B EN and a subset of the LAION-5B; the LAION 2B Multi, used for non-members, is actually a subset of the LAION-5B dataset, therefore invalidating the assumption that non-member images were never seen during model training.
\cite{japan} instead carry out extensive experiments on hyperparameters of MI attacks against stable diffusion models. The paper's primary goal is a comparison with GANs, more traditional generative networks, and hyperparameters of diffusion models, including timesteps, sampling variance, and sampling steps. The authors extensively experimented with denoising diffusion implicit models (DDIM) for the diffusion models and with a deep convolutional GAN (DCGAN) for the GAN. This research work also uses toy models trained on the CIFAR-10 and CelebA datasets. \citet{japan} implement a white-box MI attack based on the LOGAN attack \cite{logan} and a black-box one based on GAN-Leaks \cite{gan_leaks}. Results show that the white-box attacks are quite effective (AUC-ROC values between 0.552 and 0.778). However, the black-box attacks fail at distinguishing members from non-member images (AUC-ROC between 0.495 and 0.503, roughly equivalent to a toin coss).

\section{Method}
This section describes how we perform membership inference attacks on a stable diffusion model. The MI attack algorithm we propose is a pipeline of modules that can be adjusted to adapt to different threat models and application needs. In short, the algorithm determines membership as in \Cref{fig:system_schematic}:

\begin{figure*}
    \centering
    \begin{tikzpicture}

\node (f)  {$f_\theta$};
\node (t)  [below = 0 of f] {$\threatmodel$};
\node (D)  [below = 0 of t] {$\distribution$};
\node (Dt) [below = 0 of D] {$D_\text{train}$};

\node (I)      [block, label=above:Informer, block, right = 1.5 of t] {$\informer$};
\node (I_lbl)  [items, below = .25 of I] {-- mask $f_\theta$};
\node (I_lbl)  [items, right = 0 of I_lbl.south west] {-- produce};
\node (I_lbl)  [items, below right = -.25 and 0 of I_lbl.south west] {~~membership};
\node (I_lbl)  [items, below right = -.25 and 0 of I_lbl.south west] {~~dataset};
\path[->]
    (f) edge (I) 
    (t) edge (I)
    (D) edge (I)
    (Dt) edge (I);

\node (E)   [right = 5 of I] {$\encoder$};
\node (O)   [below = 0 of E.south west, draw] {$\observer$};
\node (d)   [below = 0 of E.south east, draw] {$d$};
\node (E)   [block, fit={(E) (O) (d)}, label=above:Encoder] {};
\node (E_lbl)  [items, below = 0.25 of E] {-- produce};
\node (E_lbl)  [items, right = 0 of E_lbl.south west] {~~list of};
\node (E_lbl)  [items, below right = -.2 and 0 of E_lbl.south west] {~~features $E$};
\path[->]
    (I) edge[bend left=15] 
        node[above] {$D_\text{non--member}^\text{leak}$,} 
        node[below] {$D_\text{member}^\text{leak}$}
    (E)
    (I) edge[bend right=15] node[below] {$\hat{f_\theta}, \threatmodel$} (E);

\node (C)      [block, right = 5 of E, label=above:Classifier] {$\classifier$};
\node (C_lbl)  [items, below = 0.25 of C] {-- train MI};
\node (C_lbl)  [items, below right = -.2 and 0 of C_lbl.south west] {~~binary};
\node (C_lbl)  [items, right = 0 of C_lbl.south west] {~~classifier};
\node (Out)    [right = .5 of C] {$C$};
\path[->]
    (E) edge[bend left=15] 
        node[above] {$D_\text{non--member}^\text{leak}$,} 
        node[below] {$D_\text{member}^\text{leak}$}
    (C)
    (E) edge[bend right=15]  node[below] {$E$} (C)
    (C) edge (Out);

\end{tikzpicture}
    \caption{Visual representation of the components of the MI attack system proposed in this paper. The setup is compatible with most works from the literature.}
    \label{fig:system_schematic}
\end{figure*}
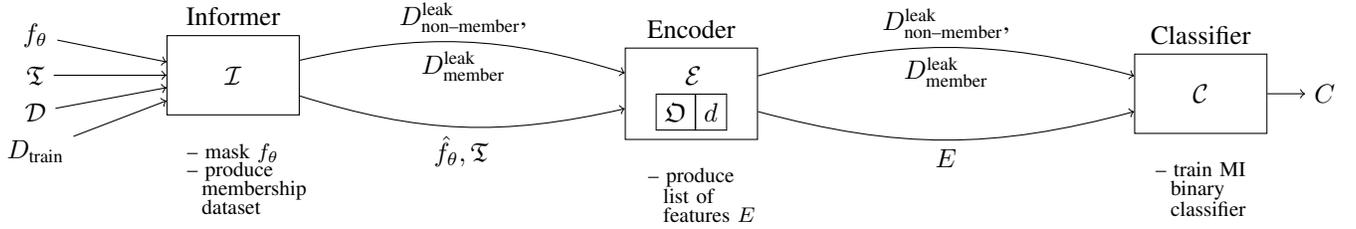

\begin{enumerate}
    \item An \textit{Informer} module determines the information that the attacker has available to carry out the membership inference. We propose two gray-box Informers with access to the victim model's intermediate diffusion steps but not the model itself. We also propose a strictly black-box Informer with access only to the samples to predict the membership of and the final outputs of the model.
    \item An \textit{Encoder} module uses the information supplied by the Informer and computes a feature vector for every sample whose membership to the training set is to be determined. This is the key component of the algorithm, which is supposed to extract information about the membership of samples.
    \item A \textit{Classifier} module processes the features supplied by the Encoder and produces a membership score for each sample. This membership score can be interpreted as the probability that the sample is a member of the training dataset; however, in practice, this is used in an ROC setting to calculate the AUC-ROC of the classifier.
\end{enumerate}

The following four subsections describe the threat model we follow and its implementation and how the three algorithmic modules can be built to maximize the membership inference attacker's power.

\subsection{Threat Model}
As mentioned in \Cref{sec:security_game}, the attacker has limited knowledge of the victim model. Traditionally, for black-box attacks, attackers are given the following three concessions:
\begin{itemize}
    \item Knowledge of the data distribution that the victim model was trained on. This is implemented by allowing the attacker to sample (with replacement) infinite points from the victim model's training data distribution. Consequently, given a discrete or continuous distribution $\distribution$, it is assumed that the victim model is trained on a subset of said distribution $D_\text{train} \subset \distribution$, subject to $\distribution / D_\text{train} \neq \varnothing$.
    \item Oracle access to the victim model. This is implemented by letting the attacker query the victim model an unlimited number of times, but having visibility only on the model's final outputs. In the context of stable diffusion, the final output is the image generated at the last diffusion step.
    \item Unlimited compute power. This is implemented by simply not restricting the amount of computation or execution time that the attacker is allowed.
\end{itemize}

Knowledge of the training data distribution with unlimited sampling and unlimited computing power has encouraged recent works on MI inference to follow the shadow model training method. Such a practice technically requires more than black-box access to the victim model, as the attacker would need to know, at least approximately, the internal architecture of the victim model and how it was trained. However, these requirements are now allowed in black-box settings with the argument that attackers targeting state-of-the-art (SOTA) systems can guess or otherwise brute-force their underlying architecture, as the number of SOTA architectures is limited, and as long as the training objective is equivalent and enough training epochs are consumed, model training does not necessarily need to follow the same procedure to yield similar accuracy as the victim model.

Most works in MI attacks that implement shadow model training techniques use either toy models, toy datasets, or both. This is because the algorithms require training many shadow models, which may be computationally unfeasible even with large-scale equipment. For instance, in the LiRA attack as proposed by \citet{lira}, the attacker trains between 32 and 256 shadow models per sample to determine its membership. A small dataset of 100 images would need 3K to over 25K models to be trained separately from scratch. Simpler attacks require training fewer shadow models, such as the label-only attack \cite{MIA_FP_60}. In the context of this research, however, a similar solution is unfeasible: training a single instance of stable diffusion costs several hundred thousand dollars (see example\footnote{\url{https://twitter.com/emostaque/status/1563870674111832066}}), making this research effort economically unrealizable. In the alternative, we leak the true membership labels of a small subset of the training data distribution to the attacker. 

We argue that the setting is equivalent for two reasons. First, the threat model assumes infinite sampling from the training data distribution, infinite computing power, and partial knowledge of the victim model architecture and training procedure, which allows an attacker to train infinite similar shadow models. The union of the shadow models allows the attacker to synthesize a training dataset of model outputs with membership labels, which in turn can be used to train a binary classifier on those labels. Given the theoretically infinite and practically large number of shadow models that the attacker can train, the distribution of the outputs of the victim model will tend to be that of the ensemble of shadow models, reducing the generalization gap of the trained membership classifier on the victim model the more shadow models there are. When the generalization gap approaches zero, the synthesized dataset becomes equivalent to the training dataset of the victim model coupled with non-training samples from the same distribution. We choose to directly release to the attacker instead of training the shadow models for obvious economic reasons.

Second, considering the vast data duplication in common large-scale vision datasets \cite{MIA_vs_SD_USENIX}, an attacker can reasonably guess at least a few training images with high confidence, therefore constructing a miniature labeled dataset of members and non-member samples. We effectively realize this procedure without manually picking samples by leaking a small subset of the training data samples to the attacker.

The following subsection describes how the black-box informer is implemented and how a gray-box variant is constructed.

\subsection{Informer Module}
\label{sec:informer}
The informer module determines the visibility of the attacker on the information about the victim model. In practice, the Informer module is a function $\informer$ that takes as input a threat model $\threatmodel$ (that can be either black-box, noted as $\blackbox$, or gray-box, noted as $\graybox$), the victim model training data distribution $\distribution$ and its training data subset $D_\text{train}$, and the victim model $f_\theta$. It then returns two key components: a masked victim model $\hat{f_\theta}$ depending on the threat model, and a partial training and testing dataset with membership labels to train the adversary MI classifier (that we note as being \textit{leaked}, to emphasize how it skips the shadow models training phase).
\begin{equation}
\begin{gathered}
    \informer\left(
        f_\theta, \threatmodel, \distribution, D_\text{train}
    \right) = 
    \left\langle 
        \hat{f_\theta}, D_\text{member}^\text{leak}, D_\text{nonmember}^\text{leak}
    \right\rangle,
    \\
    \threatmodel \in \left\{ \graybox, \blackbox \right\},
    \\
    D_\text{nonmember}^\text{leak} \subset \distribution \quad\wedge\quad D_\text{member}^\text{leak} \subseteq D_\text{train}
    \\ 
    \text{s.t.} \; D_\text{member}^\text{leak} \cap D_\text{nonmember}^\text{leak} = \varnothing
\end{gathered}
\end{equation}

The victim model can be masked in one of two ways: if the threat model is black-box, the model can be passed an input data point $(x,y) \in \mathcal{X} \times \mathcal{Y}$ (an image and text prompt pair) and optional diffusion parameters $\mathfrak{P}$, and returns only a single final image; if gray-box, the model will also return all the intermediately generated images across $T$ generation steps.

\begin{equation}
\begin{gathered}
    \hat{f_\theta} : 
    \begin{cases}
        \mathcal{X} \times \mathcal{Y} \times \mathfrak{P} \to \mathcal{X}^1 & \threatmodel = \blackbox \\
        \mathcal{X} \times \mathcal{Y} \times \mathfrak{P} \to \mathcal{X}^T & \threatmodel = \graybox \\
    \end{cases}
\end{gathered}
\end{equation}

\subsection{Encoder Module}
The encoder module $\encoder$ is the most important part of our proposed MI attack algorithm. It is responsible for generating the feature vectors that are directly supplied to and used to train the membership inference binary classifier. In short, the encoder observes how the victim stable diffusion model behaves when fed a particular sample and summarizes the observations information in a feature vector. Such a feature vector is computed by analyzing how the output of the victim model changes over time. We distinguish three ways to observe the model outputs and describe them below, along with the associated threat models. To encode information about the observation, the encoder compares pairs of images to find their similarity. We examine four similarity metrics in this section. The combination of three observation modes and four similarity metrics yields twelve versions of the encoder, whose results are thoroughly discussed in \Cref{sec:results}. \Cref{algo:encoder} formalizes the encoder process, which requires an observer $\observer$ and a similarity function $d$, discussed below.

\begin{algorithm}
    \caption{The Encoder process operates on a masked victim model $\hat{f_\theta}$ and a dataset $D$ of image-text pairs, using a similarity metric $d$ and observation mode $\observer$, to return a list of feature $E$ s.t. $|E| = |D|$, each element of length $F$.}
    \label{algo:encoder}
    \begin{algorithmic}[1]
        \Require $\hat{f_\theta}, D, \observer, d$
        \State $E \gets [\;]$
        \For {$\langle x, y \rangle \in D$}
            \State $e \gets \observer(\hat{f_\theta}, x, d)$
            \If {$|e| > 1$}
                \State $E \gets E \append e$
            \Else
                \State $E \gets E \append [e]$
            \EndIf
        \EndFor
        \State \Return $E$
    \end{algorithmic}
\end{algorithm}

\subsubsection{Observation Modes}
\label{sec:observer}
We refer to observation mode or observer as the method that the encoder uses to observe the behavior of the victim stable diffusion model. Consider a stable diffusion process where model $f$ generates an image $x_T$ from starting image $x_0 = x$ and text prompt $y$, with guidance $g$, in $T$ diffusion steps.

Three scenarios can arise depending on whether the attacker has black-box or gray-box access to the model.

\paragraph{One-Shot}\label{sec:one-shot_observer} In a purely \textit{black-box} setting, the attacker is only allowed to see the final output of the victim stable diffusion model, $x_T$. We consider the attacker to be then able only to make a single observation about the diffusion process: comparing the input image $x_0$ to $x_T$. In this case, the output of the encoder for a data point is a single number, as in \Cref{eq:one-shot}, agnostic to the distance function $d$ used.
\begin{equation}
\label{eq:one-shot}
    \threatmodel = \blackbox \wedge \observer_\text{one-shot}(\hat{f_\theta}, x, d) = d(x, \hat{f_\theta}(x))
\end{equation}

\paragraph{Progressive}\label{sec:progress_observer} If the threat model allows \textit{gray-box} access to the victim model, and therefore can see intermediate diffusion steps, we can define a first observer that analyzes how the generated image changes over time, progressing from $x_0$ to $x_T$. Such an observer is defined in \Cref{eq:progressive} below; note that the original image $x$ is not taken into consideration here to produce the encoded feature set $E$.
\begin{equation}
\label{eq:progressive}
\begin{gathered}
    \threatmodel = \graybox \wedge \observer_\text{progress}(\hat{f_\theta}, x, d) = E
    \\
    E_t = \{d(\hat{f_\theta}(x)_{t-1}, \hat{f_\theta}(x)_t), \forall t \in [2, T]\}
\end{gathered}
\end{equation}

\paragraph{Complete} If the threat model allows \textit{gray-box} access to the victim model, another approach to observing the model outputs is to study how the intermediate diffusion steps change in relation to the original image. \Cref{eq:progressive} defines this process for any distance function $d$.
\begin{equation}
\label{eq:complete}
\begin{gathered}
    \threatmodel = \graybox \wedge \observer_\text{complete}(\hat{f_\theta}, x, d) = E
    \\
    E_t = \{d(x, \hat{f_\theta}(x)_t), \forall t \in [1, T]\}
\end{gathered}
\end{equation}

\subsubsection{Similarity Metrics}
\label{sec:distance_metrics}
The observers proposed above are agnostic to the distance metric function used to compare two images. Here, we list four metrics we use throughout the experiments in \Cref{sec:evaluation}: PSNR, DSSIM, RMSE, and feature vector Euclidean distance. The first three methods require that in any pair of images to compare, the two images have the same dimensions. The diffusion process sometimes alters the dimension of the images, and we, therefore, apply a rescaling filter to scale up the smaller image to the larger image's size. In the following methods, we consider pairs of images \(x, x' \in [0, 255]^{H \times W \times 3}\).

\paragraph{PSNR}\label{sec:psnr} The Peak Signal-to-Noise Ratio (PSNR) is a commonly used image quality metric that assesses the fidelity of a processed or compressed image when compared to the original. It quantifies the ratio of the peak power of the original image to the power of the distortion, measured in terms of pixel-wise differences. Higher PSNR values indicate better image quality, as they reflect a lower level of noise or distortion in the processed image, making PSNR valuable in applications like image compression and video encoding.
\begin{equation*}
\begin{gathered}
    MSE(x, x') = \frac{\sum_{h=1}^H \sum_{w=1}^W \sum_{c=1}^3 (x_{hwc} - x'_{hwc})^2}{3HW}
    \\
    PSNR(x, x') = 10 \cdot \log_{10} \left(\frac{255^2}{MSE(x, x')}\right)
\end{gathered}
\end{equation*}

\paragraph{DSSIM} Structural Dissimilarity (DSSIM) measures the structural dissimilarity between two images. It evaluates both luminance and texture information, making it a comprehensive image quality assessment metric. DSSIM returns a value between 0 and 1, where 0 indicates identical images and higher values represent greater dissimilarity. It's particularly useful in applications where preserving the visual quality of images is critical, such as image processing, compression, and enhancement. The formula to calculate the DSSIM between two images $x$ and $x'$ is shown simplified for the context of a pair of square images below, where the channel-wise $c$ dissimilarities are calculated:
\begin{gather*}
    SSIM(x_c, x'_c) = \frac
        {(2 \mu_{x_c} \mu_{{x'}_c} + a) (2 \sigma_{x_c x'_c} + b)}
        {(\mu_{x_c}^2 + \mu_{{x'}_c}^2 + a) (\sigma_{x_c}^2 + \sigma_{x'_c}^2 + b)}
    \\
    DSSIM(x_c, x'_c) = \frac{1 - SSIM(x_c, x'_c)}{2}
\end{gather*}

Where $\mu_{x_c}, \mu_{x'_c}$ are the average pixel values of the two images on channel $c$, their variances are $\sigma_{x_c}$ and $\sigma_{x'_c}$, and the covariance is $\sigma_{x_c x'_c}$. $a, b$ are two very small variables to stabilize divisions with small denominators.    

\paragraph{RMSE} The Root Mean Square Error (RMSE) is a widely used metric for quantifying the accuracy of predictive models in regression or time series forecasting. It measures the square root of the average of the squared differences between predicted values and actual observed values, giving higher weight to larger errors. In the context of images, it quantifies the average magnitude of the pixel-wise difference between two images, with lower RMSE values indicating greater similarity between them. 
\[
    RMSE(x, x') = \sqrt{MSE(x, x')}
\]

\paragraph{Feature Vector Distance} As a fourth option, we propose to measure image similarity by first re-encoding them as visual feature vectors and then computing the Euclidean distance between the two vectors. Similar vectors suggest similar images. To extract the feature vectors, we use a pre-trained ResNet50 model\footnote{\url{https://pytorch.org/vision/main/models/generated/torchvision.models.resnet50.html}} trained to a top-1 accuracy of 80.9\% on the ImageNet dataset, from which we remove the last classification layer to expose the underlying 2048-dimensional encoder. We refer to this model as $Enc$ and the feature vector distance between two images as $Euclid$:
\[
    Euclid(x, x') = \sqrt{\sum_{i=1}^{2048} (Enc(x)_i - Enc(x')_i)^2}
\]

\subsection{Classifier Module}
The \textit{Classifier} module $\mathcal{C}$ is the final component of the MI pipeline we propose and is responsible for deciding which samples are members and which are non-members based on the feature vectors returned by the encoder. The classifier is a binary classification model whose purpose is only to best separate members from non-members, so the most critical component in the system, from an attacker's power perspective, is still the encoder module.

For the sake of completeness, every MI attack we consider uses five different classifiers, which are trained and evaluated on the same datasets. Regardless of the type, they are all defined as in \Cref{eq:classifier}, where $F$ is the size of the feature vector returned by the encoder module (see \Cref{algo:encoder} for reference).
\begin{equation}
\label{eq:classifier}
    \classifier : \mathbb{R}^F \to [0,1]
\end{equation}

The five classifiers are as follows:
\begin{enumerate}
    \item \textit{Logistic Regression} is a statistical method used for binary classification in scientific research. It models the probability of an event occurring, such as disease diagnosis or species presence, based on one or more predictor variables. Unlike linear regression, which predicts continuous outcomes, logistic regression employs the logistic function to transform linear combinations of predictors into probabilities between 0 and 1. 
    \item A \textit{Support Vector Machine} is a powerful binary classifier. SVMs excel at separating data points into two distinct classes by finding the optimal hyperplane that maximizes the margin between them. They are particularly valuable when dealing with complex, high-dimensional data, making them applicable in bioinformatics and image analysis, especially in our context of feature-encoded MI attacks.
    \item A \textit{Decision Tree} constructs a hierarchical structure of decision rules, branching based on the values of input features, ultimately producing binary outcomes. These trees are particularly valuable for their interpretability and ability to handle both categorical and continuous data.
    \item A \textit{Naive Bayes Classifier} leverages Bayes' theorem and makes a "naive" assumption that features are conditionally independent, simplifying probability calculations. These classifiers are particularly effective in handling high-dimensional data and can be trained quickly. Given their quick training and good predictive accuracy, they are widely used in machine learning research. 
    \item The \textit{K-Nearest Neighbors} (k-NN) algorithm classifies data points based on their proximity to neighboring points in a feature space. By iteratively assigning the most common class label among its k nearest neighbors, the k-NN algorithm effectively captures local data patterns and decision boundaries. Contrary to some previous models, the k-NN algorithm struggles as the dimensionality of the data increases due to the sparsity of high-dimensional clusters.
\end{enumerate}


\section{Evaluation}
\label{sec:evaluation}
This section lists the results of our MI attack against stable diffusion models. We first describe how the experiments are carried out in \Cref{sec:experiment_setup}, then in \Cref{sec:insight} show the insight that motivated this research, and finally list all results in \Cref{sec:results}.

\subsection{Experimental Setup}
\label{sec:experiment_setup}
\begin{figure}
    \centering
    \includegraphics[width=\columnwidth]{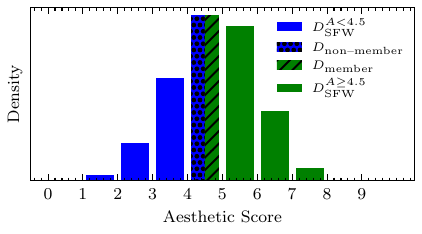}
    \caption{Distribution of the training data and how it is split into membership-labeled subsets.}
    \label{fig:dataset_split}
\end{figure}

Stable diffusion models are immensely large, and training one even once is expensive. We have therefore decided to use the single most state-of-the-art stable diffusion model throughout the experiments in this manuscript: StabilityAI's Stable Diffusion V2 \cite{SD_V2}. Stable Diffusion V2 is the first attempt at using latent diffusion models (LDMs) for image synthesis. \citet{SD_V2} demonstrate that LDMs can achieve remarkable results with significantly reduced computational requirements compared to traditional pixel-based diffusion models, thanks to the use of powerful pre-trained autoencoders in the latent space. They also highlight the addition of cross-attention layers to the model architecture, which empowers LDMs to serve as versatile generators for various conditioning inputs, resulting in state-of-the-art performance in tasks like image inpainting, scene synthesis, and super-resolution.

Stable Diffusion V2 (hereafter referred to simply as \textit{"the model"}) is trained from scratch with a subset of the LAION-5B dataset \cite{laion}, filtered for explicit pornographic material using the LAION-NSFW classifier\footnote{\url{https://github.com/LAION-AI/CLIP-based-NSFW-Detector}} with punsafe set to 0.1, and using only images with relatively high quality (Aesthetics\footnote{\url{https://github.com/christophschuhmann/improved-aesthetic-predictor}} $\geq$ 4.5). These two conditions on the LAION-5B dataset subset it unequivocally. As all the images satisfying these conditions are used for training the model, non-member samples must be fetched from elsewhere. We, therefore, propose to use a specific subset of the LAION-5B dataset and construct the membership-labeled dataset as follows as not to invalidate any assumptions made in the security game in \Cref{sec:security_game}:
\begin{enumerate}
    \item Filter out explicit material from the LAION-5B dataset with an unsafe set to 0.1. This ensures the same distribution as that used for sampling training points. This dataset is therefore $D_\text{SFW}$.
    \item Subset $D_\text{SFW}$ for samples whose Aesthetics score is between 4.5 and 5, to find $D_\text{SFW}^{4.5 \leq A < 5}$. This is, in turn, a subset of the actual training dataset of the model.
    \item Subset $D_\text{SFW}$ for samples whose Aesthetics score is between 4 and 4.5, to find $D_\text{SFW}^{4 \leq A < 4.5}$. This is a subset whose samples are very similar to those used for training, that is $D_\text{SFW}^{4 \leq A < 4.5} \sim D_\text{SFW}^{4.5 \leq A < 5}$, and yet $D_\text{SFW}^{4 \leq A < 4.5} \cap D_\text{SFW}^{4.5 \leq A < 5} = \varnothing$.
    \item Sample two datasets $D_\text{member}$ and $D_\text{non-member}$ from $D_\text{SFW}^{4.5 \leq A < 5}$ and $D_\text{SFW}^{4 \leq A < 4.5}$ respectively, each of size 1000.
\end{enumerate}

These are the membership-labeled datasets we use in the rest of this paper. \Cref{fig:dataset_split} shows the dataset selection process visually; its data refers to the official distribution statistics\footnote{\url{https://github.com/LAION-AI/laion-datasets/blob/main/laion-aesthetic.md}}.

\subsection{Manual Membership Inference}
\label{sec:insight}
\begin{figure*}
    \centering
    \includegraphics[width=\textwidth]{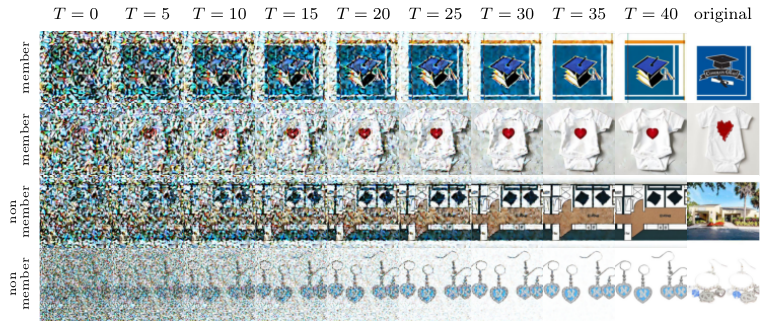}
    \caption{Examples of four images being generated, along with their original images counterparts. The top two images were part of the training set of the victim stable diffusion model, and the diffused images show strong similarity with them. The bottom two, on the other hand, were never seen during training, and in fact the generated images look quite different from the original ones.}
    \label{fig:diffusion_examples}
\end{figure*}
 
Initial experimentation with stable diffusion models led to a simple yet powerful discovery: generated images often resemble existing images that were not used in the generated diffusion sessions. This resemblance implies that the existing image was used as part of a training set for the diffusion model. To test this behavior, we chose some images from the training dataset of Stable Diffusion V2 and tried to perform diffusion using them as starting points. The results were striking: generated images would tend to reconstruct the original images. We show this tendency in \Cref{fig:diffusion_examples}, where four examples of diffusions are shown. The first two rows show images sampled from $D\text{member}$, and the bottom two are sampled from $D\text{non-member}$. Clearly, the noisy images are indistinguishable at the first step of diffusion (leftmost column). Then, images form as steps proceed (going right across columns). The samples generated from member data points show a striking similarity to their original counterparts, much more than that of non-members. The children's pajamas are nearly identical, and the graduation poster shows the same color palette and vertical and horizontal lines of the training sample, which are hardly describable with text alone. This realization is at the basis of the MI attacks proposed in this paper.


\subsection{Results}
\label{sec:results}
\begin{figure*}
    \centering
    \includegraphics[width=\textwidth]{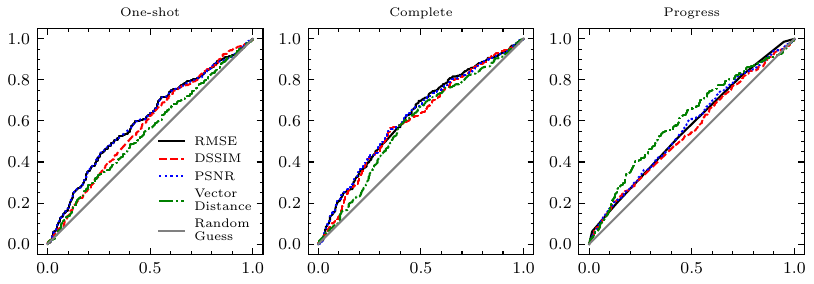}
    \caption{ROC curves relative to trained MI attack pipelines with different observers across plot, and different distance metrics within each plot. Images have not been smoothed in this experiment.}
    \label{fig:roc_linear_normal}
\end{figure*}
\begin{figure*}
    \centering
    \includegraphics[width=\textwidth]{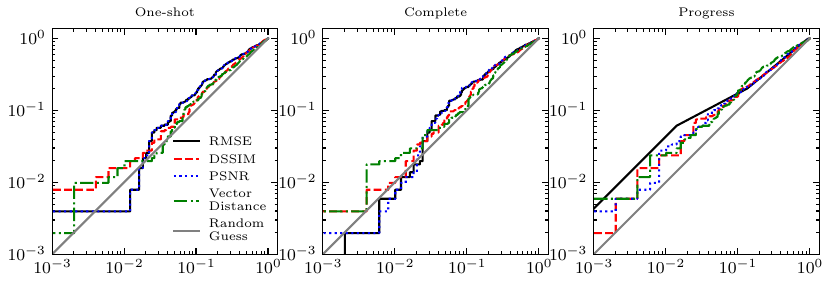}
    \caption{ROC curves relative to trained MI attack pipelines with different observers across plot, and different distance metrics within each plot. The graphs are equivalent to those of \Cref{fig:roc_linear_normal}, but use a log scale here.}
    \label{fig:roc_log_normal}
\end{figure*}
\begin{table}
\centering
\caption{AUC-ROC of the MI attack we propose, across all combinations of observation mode and distance function, with five binary classifiers for each. \textbf{Bold} values refer to the best result across binary classifiers, and \textbf{bold} and \textit{italic} to the best within the same observer class (or similarly, threat model).}
\label{tab:auc-roc_normal}
\begin{tabular}{llccc}
\toprule
 &  & One-shot & Complete & Progress \\
\midrule
\multirowcell{5}{PSNR} 
 & Logistic Regression      & \textbf{\textit{0.605}} & \textbf{\textit{0.622}} & 0.566 \\
 & Support Vector Machines  & 0.603 & 0.620 & \textbf{0.570} \\
 & Decision Trees           & 0.557 & 0.538 & 0.494 \\
 & Naive Bayes              & 0.603 & 0.620 & 0.560 \\
 & K-Nearest Neighbor       & 0.577 & 0.575 & 0.563 \\
\midrule
\multirowcell{5}{RMSE} 
 & Logistic Regression      & \textbf{0.604} & 0.621 & 0.559 \\
 & Support Vector Machines  & 0.603 & \textbf{\textit{0.622}} & 0.564 \\
 & Decision Trees           & 0.533 & 0.536 & 0.507 \\
 & Naive Bayes              & \textbf{0.604} & 0.621 & 0.553 \\
 & K-Nearest Neighbor       & 0.554 & 0.573 & \textbf{0.568} \\
\midrule
\multirowcell{5}{DSSIM} 
 & Logistic Regression      & 0.576 & \textbf{0.607} & 0.541 \\
 & Support Vector Machines  & \textbf{0.577} & 0.605 & \textbf{0.552} \\
 & Decision Trees           & 0.530 & 0.518 & 0.503 \\
 & Naive Bayes              & 0.576 & 0.593 & 0.532 \\
 & K-Nearest Neighbor       & 0.549 & 0.558 & 0.545 \\
\midrule
\multirowcell{5}{Vector\\Distance} 
 & Logistic Regression      & \textbf{0.546} & 0.567 & 0.577 \\
 & Support Vector Machines  & 0.545 & 0.570 & 0.581 \\
 & Decision Trees           & 0.511 & 0.540 & 0.531 \\
 & Naive Bayes              & 0.538 & \textbf{0.585} & \textbf{\textit{0.607}} \\
 & K-Nearest Neighbor       & 0.526 & 0.546 & 0.544 \\
\bottomrule
\end{tabular}
\end{table}

\begin{table}
\centering
\caption{AUC-ROC of the MI attack we propose, but for smoothed images with box blur of unit radius. See \Cref{tab:auc-roc_normal} for more details.}
\label{tab:auc-roc_smooth}
\begin{tabular}{llccc}
\toprule
 &  & One-shot & Complete & Progress \\
\midrule
\multirowcell{5}{PSNR} 
 & Logistic Regression      & \textbf{0.580} & 0.589 & 0.556 \\
 & Support Vector Machines  & 0.579 & 0.591 & 0.559 \\
 & Decision Trees           & 0.503 & 0.513 & 0.511 \\
 & Naive Bayes              & \textbf{0.580} & \textbf{0.600} & 0.572 \\
 & K-Nearest Neighbor       & 0.492 & 0.529 & \textbf{0.582} \\
\midrule
\multirowcell{5}{RMSE}
 & Logistic Regression      & 0.580 & 0.593 & 0.550 \\
 & Support Vector Machines  & \textbf{\textit{0.581}} & 0.597 & 0.559 \\
 & Decision Trees           & 0.517 & 0.515 & 0.513 \\
 & Naive Bayes              & 0.580 & \textbf{\textit{0.601}} & 0.560 \\
 & K-Nearest Neighbor       & 0.521 & 0.536 & \textbf{0.580} \\
\midrule
\multirowcell{5}{DSSIM} 
 & Logistic Regression      & 0.560 & 0.586 & 0.546 \\
 & Support Vector Machines  & \textbf{0.561} & \textbf{0.588} & 0.546 \\
 & Decision Trees           & 0.503 & 0.516 & 0.513 \\
 & Naive Bayes              & 0.560 & 0.581 & 0.532 \\
 & K-Nearest Neighbor       & 0.487 & 0.524 & \textbf{0.561} \\
\midrule
\multirowcell{5}{Vector\\Distance} 
 & Logistic Regression      & \textbf{0.572} & 0.567 & 0.566 \\
 & Support Vector Machines  & 0.430 & 0.574 & 0.568 \\
 & Decision Trees           & 0.499 & 0.482 & 0.522 \\
 & Naive Bayes              & 0.561 & \textbf{0.577} & \textbf{\textit{0.597}} \\
 & K-Nearest Neighbor       & 0.502 & 0.490 & 0.592 \\
\bottomrule
\end{tabular}
\end{table}

A complete evaluation of the twelve combinations of observer and distance metrics is shown in \Cref{tab:auc-roc_normal}. Results are presented as AUC-ROC, as discussed in \Cref{sec:background_evaluation}. We fit five binary classifiers on the data for each of the twelve encoders and measure their AUC-ROC on the holdout dataset as described in \Cref{sec:experiment_setup}.

Collected evidence shows a significant privacy risk: several encoders achieve and even surpass the 0.6 threshold of AUC-ROC. The highest observed performance is that of an encoder that attacks the victim model in a gray-box setting, using a \textit{complete} observer and either the \textit{PSNR} or \textit{RMSE} distance functions. For both encoders, logistic regression, and support vector machine models can best distinguish between members and non-members.

The \textit{progressive} observer, which one would expect to have similar performance to the \textit{complete} observer as they both run in gray-box settings, actually shows a consistently worse performance when stacked against the latter. It is, in fact, able to surpass the 0.6 threshold only once and is still worse than the \textit{complete} observer. Overall, while still better than randomly guessing, this method does not perform as expected, regardless of the distance metric and the trained binary classifier.

Surprisingly, the encoder with the \textit{one-shot} observer can almost match the performance of the \textit{progressive} observer while running in a strictly black-box setting. This highlights a worrying privacy threat even for commercial stable diffusion services that only expose inference APIs to their customers.

The same experiments are run after smoothing all images with box blurring of unit radius, including the input, the intermediate outputs, and the final output, and results are shown in \Cref{tab:auc-roc_smooth}. The attacker power decreased in all cases, therefore making this modification to the encoder a downgrade rather than an improvement. We do not include related ROC curves for obvious reasons.

\Cref{fig:roc_linear_normal} shows the ROC curves relative to the twelve encoeders in \Cref{tab:auc-roc_normal}. The curves use a linear scale and group encoders by observer type: one-shot, complete, and progressive. Different distance metrics are compared within the same plot for each observer type, utilizing only the binary classifier that maximizes the AUC-ROC for that encoder. The curves show a uniform distribution of the attacker's power across False Positive Rate (FPR) values, indicating that the MI attacks are not strongly skewed towards either membership or non-membership precision.

Plotting the same ROC curves on a logarithmic scale, as in \Cref{fig:roc_log_normal}, as strongly suggested by \citet{lira}, shows the MI attacker's power at low FP rates. Surprisingly, the encoder with a one-shot observer using the DSSIM distance metric is the one that archives the highest recall before finding the first false positive: almost 1\% TPR at 0\% FPR. This is a concerning data point, as an attacker with access to billions of potential training images could build a large dataset of confident training members with little to no errors.


\section{Conclusion}
This paper investigates membership inference (MI) attacks against stable diffusion computer vision models, a family of privacy-breaching attacks in machine learning models. We have chosen the widely available and state-of-the-art Stable Diffusion V2 model as a testbed for our experiments. The MI attacks we propose range from gray-box to strictly black-box and can be applied to any commercial stable diffusion service without direct access to the underlying model. Our results show that a potential attacker can determine the membership of training data samples with an accuracy over 0.6 measured with the AUC-ROC method. Additionally, attackers can determine the membership of very few samples with perfect precision, showing a 1\% recall at 0\% false positive rates. Given how image datasets nowadays are in the magnitude of billions of images \cite{laion}, this poses a severe threat to user privacy.

As computer vision stable diffusion models are constantly improving and growing a larger user base, we predict that privacy in their context will be extremely important. Research efforts should concentrate on protecting user privacy for trained models and promoting responsible data collection practices to prevent any damages caused by potential data leaks. We will continue working in this direction in the future from the dual perspective of potentially malicious attackers and of ML practitioners and service providers striving to promote their users' privacy.

\subsection{Environmentally Responsible Research}
We strongly believe in environmentally responsible AI research. We are mindful of the environmental footprint associated with the computational resources used in many AI experiments nowadays, especially when creating novel services and solutions. Throughout our research, we have made conscious choices to minimize carbon emissions. To illustrate, in our experiments, we used a single privately hosted RTX 3070 GPU card, rated at 220W, for approximately 500 hours. As the average CO2 production in Mississippi is 588 g\ce{CO2}eq per \ce{kWh}\footnote{\url{https://app.electricitymaps.com/zone/US-MIDW-MISO}}, the research contained within this manuscript generated an estimated 65 kg of \ce{CO2}. It's worth noting that, had we chosen to train a stable diffusion model from scratch even just once, it could have resulted in the emission of 10+ tons of \ce{CO2}\footnote{\url{https://github.com/Stability-AI/stablediffusion/blob/main/modelcard.md}}. This emphasizes the significance of our commitment to environmentally responsible AI research, as we actively seek to reduce the carbon footprint associated with our work while still achieving valuable results in the field of generative AI.

\subsection{Socially Responsible Research}
We are deeply committed to conducting socially responsible AI research, especially in the context of exploring privacy in generative AI. As AI researchers, we recognize the profound impact that AI can have on individuals and society. As we promote privacy-centric AI and ML, we ensure that our research follows ethical AI practices and safeguards user privacy. We firmly believe that the responsible development of AI is vital to prevent the misuse of technology and protect individuals from potential privacy infringements, both intentional and not.

\bibliographystyle{IEEEtranN}
\bibliography{sources}

\end{document}